%% file: main.tex
\documentclass[runningheads]{llncs}
\usepackage{graphicx}

\usepackage{tikz}
\usepackage{comment}
\usepackage{amsmath,amssymb} 
\usepackage{color}

\usepackage[accsupp]{axessibility}  

\usepackage[ruled,vlined,linesnumbered]{algorithm2e}
\SetKwComment{Comment}{$\triangleright$\ }{}
\SetKwComment{CommentL}{$\/\/$}{}
\SetKwFunction{FMain}{Main}
\usepackage{tikz}
\usetikzlibrary{fit,calc}

\usepackage{bbding}
\usepackage{pifont}
\usepackage{wasysym}
\usepackage{amssymb}

\usepackage{multirow}
\usepackage{booktabs}
\usepackage{float}
\usepackage{wrapfig}

\newcommand*{\tikzmka}[1]{\tikz[remember picture,overlay,] \node (#1) {};\ignorespaces}
\newcommand{\boxita}[1]{\tikz[remember picture,overlay]{\node[yshift=3pt,fill=#1,opacity=.25,fit={(A)($(B)+(0.118\linewidth,.1\baselineskip)$)}] {};}\ignorespaces}

\newcommand*{\tikzmkc}[1]{\tikz[remember picture,overlay,] \node (#1) {};\ignorespaces}
\newcommand{\boxitc}[1]{\tikz[remember picture,overlay]{\node[yshift=3pt,fill=#1,opacity=.25,fit={(A)($(B)+(0.118\linewidth,.1\baselineskip)$)}] {};}\ignorespaces}
\colorlet{pink}{red!40}
\colorlet{cyan}{cyan!60}
\colorlet{orange}{orange!80}
\newcommand{\bp}[1]{\textcolor{black}{#1}}
\newcommand{\hr}[1]{\textcolor{black}{#1}}

\begin{document}
\pagestyle{headings}
\mainmatter
\def\ECCVSubNumber{6312}  

\title{SuperTickets: Drawing Task-Agnostic \\Lottery Tickets from Supernets via Jointly \\Architecture Searching and Parameter Pruning} 

\titlerunning{SuperTickets}
%
\author{Haoran You\inst{1}\thanks{Work done while interning at Baidu USA; Correspondence to Baopu Li and Yingyan Lin} \and
Baopu Li\inst{3} \and
Zhanyi Sun\inst{2} \and
Xu Ouyang\inst{2} \and
Yingyan Lin\inst{1}}
\authorrunning{H. You et al.}
%
\institute{
$^1$Georgia Institute of Technology \quad
$^2$Rice University \quad
$^3$Oracle Health and AI \\
\email{\{hyou37, celine.lin\}@gatech.edu}
\email{\{zs19,xo2\}@rice.edu},  \email{baopu.li@oracle.com}
}
\maketitle

\begin{abstract}
Neural architecture search (NAS) has demonstrated amazing success in searching for efficient deep neural networks (DNNs) from a given supernet.
In parallel, lottery ticket hypothesis has shown that DNNs contain small subnetworks that can be trained from scratch to achieve a comparable or even higher accuracy than the original DNNs.
As such, it is currently a common practice to develop efficient DNNs via a pipeline of first search and then prune. Nevertheless, doing so often requires a tedious and costly process of search-train-prune-retrain and thus prohibitive computational cost.  
In this paper, we discover for the first time that both efficient DNNs and their lottery subnetworks (i.e., lottery tickets) can be directly identified from a supernet, which we term as \textbf{SuperTickets}, via a two-in-one training scheme with jointly architecture searching and parameter pruning.
Moreover, we develop a progressive 
and unified SuperTickets identification
strategy that allows the connectivity of subnetworks to change during supernet training, achieving better accuracy and efficiency trade-offs than conventional sparse training.
Finally, we evaluate whether such identified SuperTickets drawn from one task can transfer well to other tasks, validating their potential of simultaneously handling multiple tasks.
Extensive experiments and ablation studies on three tasks and four benchmark datasets validate that our proposed SuperTickets achieve boosted accuracy and efficiency trade-offs than both \bp{typical} NAS and pruning pipelines, regardless of \bp{having retraining or not}.
Codes and pretrained models are available at \url{\textcolor{blue}{https://github.com/RICE-EIC/SuperTickets}}.

\end{abstract}

\input{sections/0-Introduction}
\input{sections/1-Related-work}
\input{sections/2-Methods}
\input{sections/3-Experiments}
\input{sections/4-Conclusion}

\clearpage
%
%
\bibliographystyle{splncs04}
\bibliography{eccv2022}

\input{sections/5-Appendix}

\end{document}

%% file: sections/0-Introduction.tex
\section{Introduction}

While deep neural networks (DNNs) have achieved unprecedented performance in 
various
tasks and applications like classification, segmentation, and detection~\cite{HR-NAS}, their prohibitive training and inference costs limit \bp{their deployment on} resource-constrained devices for more pervasive intelligence.
For example, one forward pass of the ResNet50 \cite{he2016deep} requires 4 GFLOPs (FLOPs: floating point operations) and its training requires $10^{18}$ FLOPs \cite{You2020Drawing}.
To close the aforementioned gap, extensive attempts have been made to compress DNNs from either macro-architecture (e.g.,  NAS~\cite{tan2019mnasnet,wu2019fbnet,HR-NAS}) or fine-grained parameter (e.g., network pruning~\cite{han2015deep,frankle2018the}) levels.
A commonly adopted DNN compression pipeline following a coarse-to-fine principle is to first automatically search efficient and powerful DNN architectures from a larger supernet and then prune the searched DNNs via costly train-prune-retrain process \cite{feng2021edge,ding2022nap,li2021npas} to derive smaller and sparser subnetworks with a comparable or degraded accuracy but largely reduced inference costs.
However, such pipeline requires a tedious search-train-prune-retrain process and thus still prohibitive training costs.

To address the above limitation for simplifying the pipeline and further improve the accuracy-efficiency trade-offs of the identified networks, we advocate a \textbf{two-in-one training} framework for simultaneously identifying both efficient DNNs and their lottery subnetworks via jointly architecture searching and parameter pruning.
We term the identified small subnetworks as \textbf{SuperTickets} if they achieve comparable or even superior accuracy-efficiency trade-offs than previously adopted search-then-prune baselines, 
because they are drawn from supernets and represent both coarse-grained DNN architectures and fine-grained DNN subnetworks.
We make non-trivial efforts to explore and validate the potential of SuperTickets by answering three key questions:
\textit{(1) whether such SuperTickets can be directly found from a supernet via two-in-one training? If yes, then (2) how to effectively identify such SuperTickets? and (3) can SuperTickets found from one task/dataset transfer to another, i.e., have the potential to handle different tasks/datasets?}
%
To the best of our knowledge, this is the first \bp{attempt} taken towards identifying both DNN architectures and their corresponding lottery ticket subnetworks through a unified two-in-one training scheme.
Our contributions can be summarized as follows:

\begin{itemize}
    \setlength{\itemsep}{0pt}
    \setlength{\parsep}{0pt}
    
    \item [$\bullet$] We \textbf{for the first time} discover that efficient DNN architectures and their lottery subnetworks, i.e., SuperTickets, can be simultaneously identified from a supernet 
    leading to superior accuracy-efficiency trade-offs.
    
    \item [$\bullet$] We develop an \bp{unified progressive} identification strategy to effectively find the SuperTickets via a two-in-one training scheme which allows the subnetworks to iteratively reactivate the pruned connections during training, offering better performance than conventional sparse training. Notably, our identified SuperTickets \textit{without retraining} already outperform previously adopted first-search-then-prune baselines, and thus can be directly deployed.
    
    \item [$\bullet$] We validate the transferability of identified SuperTickets across different tasks/datasets, and conduct extensive experiments to compare the proposed SuperTickets with those from existing search-then-prune baselines, \bp{ typical} NAS techniques, and pruning works. Results on three tasks and four datasets 
    demonstrate the consistently superior accuracy-efficiency trade-offs and the promising transferability for handling different tasks offered by SuperTickets.
\end{itemize}


%



%% file: sections/1-Related-work.tex
\section{Related Works}
\textbf{Neural Architecture Search (NAS).}
NAS has achieved an amazing success in automating the design of efficient DNN architectures and boosting accuracy-efficiency trade-offs \cite{zoph2018learning,tan2019efficientnet,howard2019searching}. 
To search for task-specific DNNs, early works \cite{tan2019efficientnet,tan2019mnasnet,howard2019searching} adopt reinforcement learning based methods that require a prohibitive search time and computing resources,
while recent works \cite{liu2018darts,wu2019fbnet,wan2020fbnetv2,nas_cvpr2021} update both the 
weights and architectures during supernet training via differentiable search that can greatly improve the search efficiency as compared to prior NAS works.
More recently, some works adopt one-shot NAS \cite{guo2020single,cai2019once,yu2020bignas,wang2021alphanet} to decouple the architecture search from supernet training. 
Such methods are generally applicable to search for efficient CNNs \cite{guo2020single,bender2018understanding} or Transformers \cite{wang2020hat,AutoFormer,VITAS} for solving both vision and language tasks.
To search for multi-task DNNs, recently emerging works like HR-NAS \cite{HR-NAS} and FBNetv5 \cite{wu2021fbnetv5} advocate supernet designs with multi-resolution branches so as to accommodate both image classification and other dense prediction tasks that require high-resolution representations. In this work, we propose to directly search for not only efficient DNNs but also their lottery subnetworks from supernets 
to achieve better accuracy-efficiency trade-offs while being able to handle different tasks.

\textbf{Lottery Ticket Hypothesis (LTH).}
Frankle et al. \cite{frankle2018the,frankle2020linear} showed that winning tickets (i.e., small subnetworks) exist in randomly initialized dense networks, which can be retrained to restore a comparable or even better accuracy than their dense network counterparts. This finding has inspired lots of research directions as it implies the potential of sparse subnetworks.
For efficient training, You et al. \cite{You2020Drawing} consistently find winning tickets at early training stages, largely reducing DNNs' training costs.
Such finding has been extended to language models (e.g., BERT) \cite{chen2020earlybert}, generative models (e.g., GAN) \cite{mukund2020winning}, and graph neural networks \cite{you2021early};
Zhang et al. \cite{zhang2021efficient} recognize winning tickets more efficiently by training with only a specially selected subset of data; and
Ramanujan et al. \cite{ramanujan2020s} further identify winning tickets directly from random initialization that perform well even without retraining.
In contrast, our goal is to simultaneously find both efficient DNNs and their lottery subnetworks from supernets, beyond the scope of sparse training or drawing winning tickets from dense DNN models.

\textbf{Task-Agnostic DNNs Design.}
To facilitate designing DNNs for different tasks, recent works \cite{liu2021swin,howard2019searching,wang2020deep} propose to design general architecture backbones for various computer vision tasks.
For example, 
HR-Net \cite{wang2020deep} maintains high-resolution representations through the whole network for supporting dense prediction tasks, instead of connecting high-to-low resolution convolutions in series like ResNet or VGGNet;
Swin-Transformer \cite{liu2021swin} adopts hierarchical vision transformers to serve as a general-purpose backbone that is compatible with a broad range of vision tasks;
ViLBERT \cite{lu2019vilbert,lu202012} proposes a multi-modal two-stream model to learn task-agnostic joint representations of both image and language;
Data2vec \cite{baevski2022data2vec} designs a general framework for self-supervised learning in speech, vision and language.
%
Moreover, recent works \cite{HR-NAS,wu2021fbnetv5,xu2021bert} also leverage NAS to automatically search for task-agnostic and efficient DNNs from hand-crafted supernets.
In this work, we aim to identify task-agnostic SuperTickets that achieve better accuracy-efficiency trade-offs.

%% file: sections/2-Methods.tex
\section{The Proposed SuperTickets Method}
In this section, we address the three key questions of SuperTickets. 
First, we develop a two-in-one training scheme to validate our hypothesis that SuperTickets exist and can be found directly from a supernet. 
Second, we further explore more effective SuperTickets identification strategies via iterative neuron reactivation and progressive pruning, largely boosting the accuracy-efficiency trade-offs.
Third, we evaluate the transferability of the identified SuperTickets across different datasets or tasks, validating their potential of being task-agnostic.

\subsection{Do SuperTickets Exist in Supernets?}

\textbf{SuperTickets Hypothesis.} We hypothesize that both efficient DNN architectures and their lottery subnetworks can be directly identified from a supernet, and term these subnetworks as \textbf{SuperTickets} if they achieve on par or even better accuracy-efficiency trade-offs than those from first-search-then-prune counterparts.
Considering a supernet $f(x; \theta_{S})$, various DNN architectures $a$ are sampled from it whose weights are represented by $\theta_S(a)$, then we can define SuperTickets as $f(x; m \odot \theta_S(a))$, where $m \!\in\! \{0, 1\}$ is a mask to indicate the pruned and unpruned connections in searched DNNs.
The SuperTickets Hypothesis implies that jointly optimizing DNN architectures $a$ and corresponding sparse masks $m$ works better, i.e., resulting in superior accuracy-efficiency trade-offs, than sequentially optimizing them.

\textbf{Experiment Settings.}
To perform experiments for exploring whether SuperTickets generally exist, we need (1) a suitable supernet taking both classical efficient building blocks and task-agnostic DNN design principles into consideration and (2) corresponding tasks, datasets, and metrics. 
We elaborate our settings below.
\underline{NAS and Supernets:} We consider a multi-branch search space containing both efficient convolution and attention building blocks following one state-of-the-art (SOTA) work  \bp{of HR-NAS} ~\cite{HR-NAS}, whose unique hierarchical multi-resolution search space for handling multiple vision tasks \bp{stands out} compared to others. 
In general, it contains two paths: MixConv \cite{tan2019mixconv} and lightweight Transformer for extracting both local and global context information. 
Both the number of convolutional
channels with various kernel sizes and the number of tokens in the Transformer are searchable parameters.
\underline{Tasks, Datasets, and Metrics:} We consider semantic segmentation on Cityscapes \cite{cordts2016cityscapes} and human pose estimation on COCO keypoint \cite{lin2014microsoft} as two representative tasks \bp{for illustrative purposes}.
For Cityscapes, 
the mean Intersection over Union (mIoU), mean Accuracy (mAcc), and overall Accuracy (aAcc) are evaluation metrics.
For COCO keypoint, we train the model using input size 256$\times$192, \bp{an} initial learning rate of 1e-3, a batch size of 384 for 210 epochs. The average precision (AP), recall scores (AR), AP$^M$ and AP$^L$ for medium or large objects are evaluation metrics. All experiments \bp{are} run on  Tesla \bp{V100*8} GPUs.

\begin{figure}[!t]
\centering
\begin{algorithm}[H]
\SetAlgoLined
\KwIn{The supernet weights $\theta_S$, drop threshold $\epsilon$, and pruning ratio $p$;} 
\KwOut{Efficient DNNs and their lottery subnetworks $f(x; m \odot \theta_S(a))$.}
\While{$t$ (epoch) $< t_{max}$ }{

$t = t+1$;

Update weights $\theta_S$ and importance factor $r$ using SGD training;

\uIf(\Comment*[f]{Search for DNNs}){$t\mod t_s = 0$}{

    Remove search units whose importance factors $r < \epsilon$; 
    
    Recalibrate the running statistics of BN layers to obtain subnet $a$;
    
    \CommentL{\textcolor{purple}{\textit{// If enabling the iterative reactivation technique}}}
    
    \tikzmkc{A} \textcolor{black}{Reactivate the gradients of pruned weights;} \tikzmkc{B} \boxitc{orange}

}

\ElseIf(\Comment*[f]{Prune for subnetworks}){$t\mod t_p = 0$}{

    \CommentL{\textcolor{blue}{\textit{// If enabling the progressive pruning technique}}}
    
    \tikzmka{A} \textcolor{black}{Redefine the pruning ratio as $\min\{p, 10\% \times \lfloor t / t_p \rfloor \}$;} \tikzmka{B} \boxita{cyan}
    
    Perform magnitude-based pruning towards the target ratio;
    
    Keep the sparse mask $m_t$ and disable pruned weights' gradients;

}
}


\Return $f(x; m_t \odot \theta_S(a))$\Comment*[r]{\textcolor{purple}{SuperTickets}}
\caption{Two-in-One Framework for Identifying SuperTickets.}
\label{alg:two-in-one}

\end{algorithm}
\end{figure}

\begin{wrapfigure}{r}{0.6\textwidth}
    \centering
    \includegraphics[width=\linewidth]{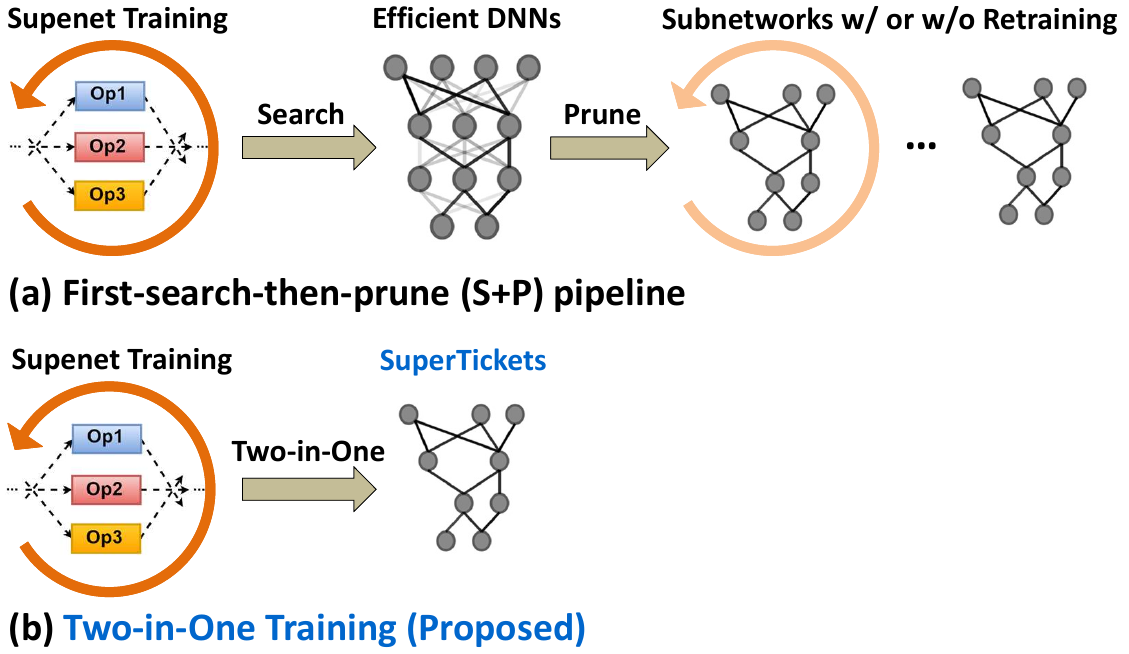}
     \caption{Illustrating first-search-then-prune (S+P) vs. our two-in-One training.}
    \label{fig:comp}
\end{wrapfigure}

\textbf{Two-in-One Training.} To validate the SuperTickets hypothesis, we propose a two-in-one training algorithm that simultaneously searches and prunes during supernet training of NAS.
As shown in Alg. \ref{alg:two-in-one} and Fig. \ref{fig:comp}, for searching for efficient DNNs, we adopt a progressive shrinking NAS by gradually removing unimportant search units that can be either convolutional channels or Transformer tokens. 
\bp{After every $p_s$ training epochs, we will detect and remove the unimportant search units once their corresponding importance factors $r$ (i.e., the scales in Batch Normalization (BN) layers) are less than a predefined drop threshold $\epsilon$.
Note that $r$ can be jointly learned with supernet weights, such removing will not affect the remaining search units since channels in depth-wise convolutions are independent among each other, as also validated by \cite{HR-NAS,mei2019atomnas}.}
In addition, we follow network slimming \cite{liu2017learning} to add a $l_1$ penalty as a regularization term for polarizing the importance factors to ease the detection of unimportant units.
After removing them, the running statistics in BN layers \bp{are recalibrated in order to match the searched DNN architecture $a$ for avoiding covariate shift \cite{ioffe2015batch,you2021test}.
}
For pruning of searched DNNs, we perform magnitude-based pruning towards the given pruning ratio per $t_p$ epochs, the generated spare mask $m_t$ will be kept so as to disable the gradients flow of the pruned weights during the following training.
Note that we do not incorporate the iterative reactivation and progressive pruning techniques (highlighted with colors/shadows in Alg. \ref{alg:two-in-one}, \bp{which will be elaborated later)} as for now. Such vanilla two-in-one training algorithm can be regarded as the first step towards answering the puzzle whether SuperTickets generally exist.

\begin{figure}[t]
    \centering
    \includegraphics[width=\linewidth]{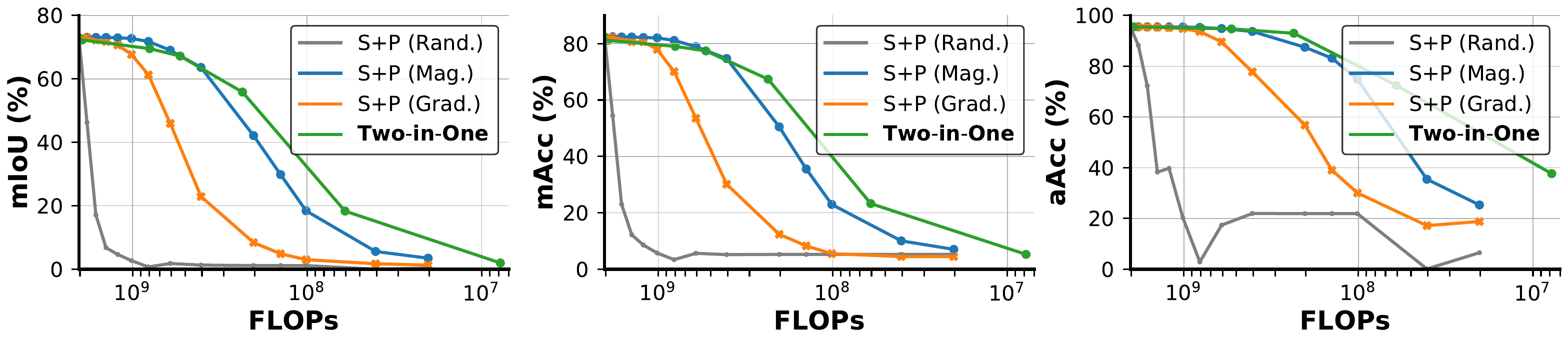}
    \caption{Comparing the mIoU, mAcc, aAcc and inference FLOPs of the resulting networks from the proposed two-in-one training and first-search-then-prune (S+P) baselines on semantic segmentation task and Cityscapes dataset, where Rand., Mag., and Grad. represent random, magnitude, and graident-based pruning, respectively. Note that each method has a series of points for representing different pruning ratios ranging from 10\% to 98\%. All accuracies are averaged over three runs.}
    \label{fig:two-in-one_cityscapes}
\end{figure}

\begin{figure}[t]
    \centering
    \includegraphics[width=\linewidth]{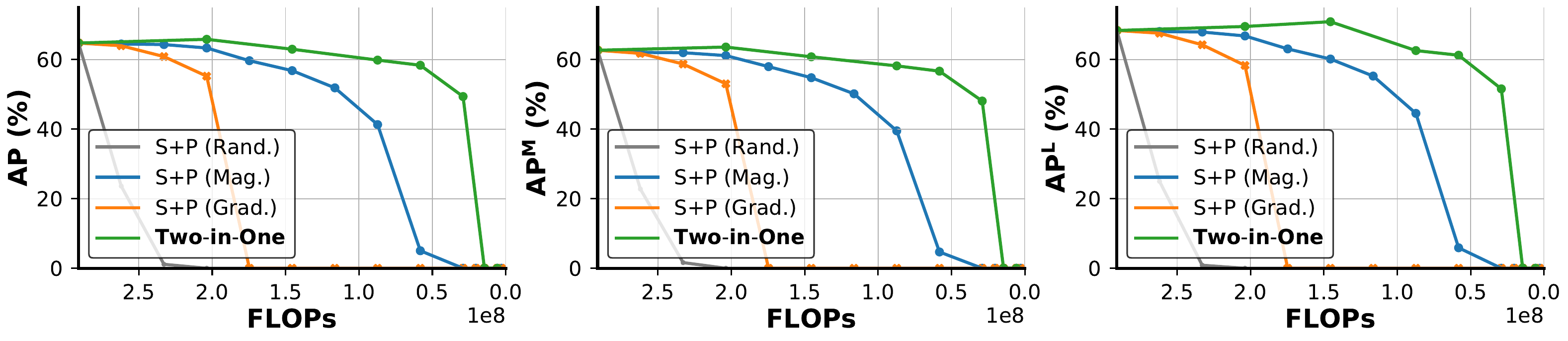}
    \caption{Comparing the AP, AP$^M$, AP$^L$ and inference FLOPs of the resulting networks from the proposed two-in-one training and baselines on human pose estimation task and COCO keypoint dataset. Each method has a series of points for representing different pruning ratios ranging from 10\% to 98\%. All accuracies are averaged over three runs.}
    \label{fig:two-in-one_coco}
\end{figure}

\textbf{Existence of SuperTickets.} We compare the proposed two-in-one training with first-search-then-prune (S+P) baselines and report the results on Cityscapes and COCO keypoint at Fig. \ref{fig:two-in-one_cityscapes} and Fig. \ref{fig:two-in-one_coco}, respectively.
We see that the proposed two-in-one training consistently generates comparable or even better accuracy-efficiency trade-offs as compared to S+P with various pruning criteria (random, magnitude, and gradient) \bp{since our methods demonstrate much better performance of segmentation or human pose estimation under different FLOPs reductions as shown in the above two figures,} indicating that SuperTickets generally exist in a supernet and have great potential to outperform the commonly adopted approaches, i.e., sequentially optimizing DNN architectures and sparse masks.

\subsection{How to More Effectively Identify SuperTickets?}
\label{sec:identifier}

We have validated the existence of SuperTickets, the natural next question is how to more effectively identify them. To this end, we propose two techniques that can be seamlessly incorporated into the two-in-one training framework to more effectively identify SuperTickets and further boost their achievable performance.


\begin{table}[t]
    \setlength{\tabcolsep}{4pt}
    \centering
    \caption{Breakdown analysis of the proposed SuperTickets identification strategy. We report the performance of found subnetworks under 90\%/80\% sparsity on two datasets.}
    \resizebox{\linewidth}{!}{
        \begin{tabular}{c|ccccc|ccc|cccc}
        \hline
        \multirow{2}[1]{*}{\textbf{Methods}} & \multirow{2}[1]{*}{\textbf{2-in-1}} & \multirow{2}[1]{*}{\textbf{PP}} & \multirow{2}[1]{*}{\textbf{IR-P}} & \multirow{2}[1]{*}{\textbf{IR-S}} & \multirow{2}[1]{*}{\textbf{Retrain}} & \multicolumn{3}{c|}{\textbf{Cityscapes}} & \multicolumn{4}{c}{\textbf{COCO Keypoint}} \\
        \cline{7-13}  &   &   &   &   &   & \textbf{mIoU} & \textbf{mAcc} & \textbf{aAcc} & \multicolumn{1}{c}{\textbf{AP}} & \multicolumn{1}{c}{\textbf{AP$^M$}} & \multicolumn{1}{c}{\textbf{AP$^L$}} & \multicolumn{1}{c}{\textbf{AR}} \\
        \hline
        \hline
        S+P (Mag.) &  &   &   &   &   & 42.12 & 50.49 & 87.45 & 5.04 & 4.69 & 5.89 & 10.67 \\
        S+P (Mag.) &   &   &   &   & \Checkmark & 51.03 & 59.61 & 90.88 & 48.63 & 46.82 & 51.74 & 53.38 \\
        \hline
        Ours & \Checkmark &   &   &   &   & 55.84 & 67.38 & 92.97 & 58.38 & 56.68 & 61.26 & 62.23 \\
        Ours & \Checkmark & \Checkmark &   &   &   & 63.89 & 73.56 & 94.17 & 60.14 & 57.93 & 63.70 & 63.79 \\
        Ours & \Checkmark & \Checkmark & \Checkmark &   &   & 45.73 & 55.52 & 89.36 & 5.48 & 7.43 & 4.36 & 10.85 \\
        Ours & \Checkmark & \Checkmark &   & \Checkmark &   & \textbf{66.61} & \textbf{76.30} & \textbf{94.63} & \textbf{61.02} & \textbf{58.80} & \textbf{64.64} & \textbf{64.78} \\
        Ours & \Checkmark & \Checkmark &   & \Checkmark & \Checkmark & \textbf{67.17} & \textbf{77.03} & \textbf{94.73} & \textbf{61.48} & \textbf{59.30} & \textbf{65.19} & \textbf{65.20} \\
        \hline
        \end{tabular}%
    }
    \label{tab:two_in_one_breakdown}
\end{table}

\textbf{Progressive Pruning (PP).} Although simultaneously searching and pruning during supernet training enables the opportunity of cooperation between coarse-grained search units removal and fine-grained weights pruning, \bp{i.e., NAS} helps to refine the pruned networks as a compensation by removing over-pruned units for avoiding bottlenecked layers,
we find that over-pruning at the early training stages inevitably hurts the networks' generalizability, and further propose a progressive pruning (PP) techniques to overcome this shortcoming.
As highlighted in the \textcolor{cyan}{\textbf{cyan}} part of Alg. \ref{alg:two-in-one}, the pruning ratio is defined as $\min \{p, 10\%\times \lfloor t / t_p\rfloor\}$, which means that the network sparsity will gradually increase from 10\% to the target ratio $p$, by 10\% per $t_p$ epochs.
The PP technique helps to effectively avoid over-pruning at early training stages and thus largely boosts the final performance.
As demonstrated in Table \ref{tab:two_in_one_breakdown}, two-in-one training with PP achieves 8.05\%/6.18\%/1.2\% mIoU/mAcc/aAcc and \hr{1.76\%/1.25\%/2.44\%/1.56\%} AP/AP$^M$/AP$^L$/AR improvements on Cityscapes and COCO keypoint datasets, respectively, as compared to the vanilla two-in-one training under 90\% sparsity.

\textbf{Iterative Reactivation (IR).} Another problem in the two-in-one framework is that the pruned weights will never \bp{get gradients updates} throughout the remaining training. To further boost the performance, we design \bp{an} iterative reactivation (IR) strategy to facilitate the effective SuperTickets identification by allowing the connectivity of subnetworks to change during supernet training.
Specifically, we reactivate the gradients of pruned weights as highlighted in the \textcolor{orange}{\textbf{orange}} part of Alg. \ref{alg:two-in-one}.
Note that we reactivate during searching instead of right after pruning, based on a hypothesis that sparse training is also essential to the two-in-one training framework. In practice, the pruning interval $p_t$ is different from the searching interval $p_s$ in order to allow a period of sparse training.
To validate the hypothesis, we design two variants: IR-S and IR-P that reactivate pruned weights' gradients during searching and pruning, respectively, and show the comparisons in Table \ref{tab:two_in_one_breakdown}.
We observe that: (1) IR-P leads to even worse accuracy than vanilla two-in-one training, validating that sparse training is essential; (2) IR-S further leads to 2.72\%/2.74\%/0.46\% mIoU/mAcc/aAcc and \hr{0.88\%/0.87\%/0.94\%/0.99\%} AP/AP$^M$/ AP$^L$/AR improvements on Cityscapes and COCO keypoint, respectively, on top of two-in-one training with PP.



\textbf{SuperTickets w/ or w/o Retraining.} Since the supernet training, architecture search, and weight pruning are conducted in an unified end-to-end manner, \hr{the resulting SuperTickets can be deployed directly without retraining, achieving better accuracy-efficiency trade-offs than S+P baselines (even with retraining) as indicated by Table \ref{tab:two_in_one_breakdown}.}
To investigate whether retraining can further boost the performance, we retrain the found SuperTickets for another 50 epochs and report the results at Table \ref{tab:two_in_one_breakdown}. We see that retraining further leads to 0.56\%/0.73\%/0.10\% mIoU/mAcc/aAcc and \hr{0.46\%/0.50\%/0.55\%/0.42\%} AP/AP$^M$/ AP$^L$/AR 
improvements on Cityscapes and COCO keypoint datasets, respectively.

\subsection{Can the Identified SuperTickets Transfer?} \label{sec:transfer}

To validate the potential of identified SuperTickets for handling different tasks and datasets, we provide empirical experiments and analysis as follows. Note that we adjust the final classifier to match target datasets during transfer learning.

\begin{wraptable}{r}{0.53\textwidth}
    \setlength{\tabcolsep}{3pt}
    \centering
    \caption{\bp{Supertickets transfer validation tests under 90\% sparsity.}}
    \resizebox{0.53\textwidth}{!}{
        \begin{tabular}{l|cc|ccc}
        \hline
        \multirow{2}[1]{*}{\textbf{Methods}} & \multirow{2}[1]{*}{\textbf{Params}} & \multirow{2}[1]{*}{\textbf{FLOPs}} & \multicolumn{3}{c}{\textbf{Cityscapes}} \\
        \cline{4-6}  &   &   & \textbf{mIoU} & \textbf{mAcc} & \textbf{aAcc} \\
        \hline
        \hline
        S+P (Grad.) & 0.13M & 203M & 8.41 & 12.39 & 56.77  \\
        S+P (Mag.) & 0.13M & 203M & 42.12 & 50.49 & 87.45 \\
        S+P (Mag.) w/ RT & 0.13M & 203M & 60.76 & 70.40 & 93.38 \\
        ADE20K Tickets & 0.20M & 247M & \textbf{62.91} & \textbf{73.32} & \textbf{93.82}  \\
        ImageNet Tickets & 0.18M & 294M & 61.64 & 71.78 & 93.75 \\
        \hline
        \hline
        \multirow{2}[1]{*}{\textbf{Methods}} & \multirow{2}[1]{*}{\textbf{Params}} & \multirow{2}[1]{*}{\textbf{FLOPs}} & \multicolumn{3}{c}{\textbf{ ADE20K}} \\
        \cline{4-6}  &   &   & \textbf{mIoU} & \textbf{mAcc} & \textbf{aAcc} \\
        \hline
        \hline
        S+P (Grad.) & 0.11M & 154M & 0.79 & 1.50 & 25.58  \\
        S+P (Mag.)  & 0.11M & 154M & 3.37 & 4.70 & 39.47 \\
        Cityscapes Tickets & 0.13M & 119M & 20.83 & 29.95 & 69.00  \\
        ImageNet Tickets & 0.21M & 189M & \textbf{22.42} & \textbf{31.87} & \textbf{70.21} \\
        \hline
        \end{tabular}%
    }
    \label{tab:transfer}
\end{wraptable}

\textbf{SuperTickets Transferring Among Datasets.} We first test the transferability of the identified SuperTickets among different datasets within the same task, i.e., Cityscapes and ADE20K as two representatives in the semantic segmentation task.
Table \ref{tab:transfer} shows that SuperTickets identified from one dataset can transfer to another dataset while leading to comparable or even better performance than 
S+P baselines with (denoted as ``w/ RT'') or without retraining (by default).
For example, when tested on Cityscapes, SuperTickets identified from ADE20K after fine-tuning lead to 2.2\% and 20.8\% higher mIoU than S+P (Mag.) w/ and w/o RT baselines which are directly trained on target Cityscapes dataset. Likewise, the SuperTickets \bp{transferred} from Cityscapes to ADE20K also outperform baselines on target dataset.

\textbf{SuperTickets Transferring Among Tasks.} To further investigate whether the identified SuperTickets can transfer among different tasks.
We consider to transfer SuperTickets's feature extraction modules identified from ImageNet on classification task to Cityscapes and ADE20K on segmentation tasks, where the dense prediction heads and final classifier are still inherited from the target datasets. The results are presented \bp{in the last row of the two sub-tables} in Table \ref{tab:transfer}. We observe that such transferred networks still perform well on downstream tasks. Sometimes, it even achieves better performance than transferring within one task, e.g., ImageNet $\rightarrow$ ADE20K works better (1.6\% higher mIoU) than Cityscapes $\rightarrow$ ADE20K.
We supply more experiments on various pruning ratios in Sec. \ref{sec:transfer_ablation}.

%% file: sections/3-Experiments.tex
\section{Experiment Results}

\subsection{Experiment Setting}
\textbf{Tasks, Datasets, and Supernets.} 
\underline{Tasks and Datasets.} 
We consider four benchmark datasets and three representative vision tasks to demonstrate the effectiveness of SuperTickets, including image classification on ImageNet \cite{deng2009imagenet} dataset with 1.2 million training images and 50K validation images; semantic segmentation on Cityscapes \cite{cordts2016cityscapes} and ADE20K \cite{zhou2017scene} datasets with 2975/500/1525 and 20K/2K/3K images for training, validation, and testing, respectively; human pose estimation on COCO keypoint \cite{lin2014microsoft} dataset with 57K images and 150K person instances for training, and 5K images for validation.
These selected datasets require different receptive fields and global/local contexts, manifesting themselves as proper test-beds for SuperTickets on multiple tasks.
\underline{Supernets.} 
For all experiments, we adopt the same supernet as HR-NAS \cite{HR-NAS} thanks to the task-agnostic multi-resolution supernet design. It begins with two 3$\times$3 convolutions with stride 2, which is followed by five parallel modules to gradually divide it into four branches of decreasing resolutions, the learned features from all branches are then merged together for classification or dense prediction.

\textbf{Search and Training Settings.}
For training supernets on ImageNet, we adopt a RMSProp optimizer with 0.9 momentum and 1e-5 weight decay, exponential moving average (EMA) with 0.9999 decay, and exponential learning rate decay with an initial learning rate of 0.016 and 256 batch size for 350 epochs.
For Cityscapes and ADE20K, we use an AdamW optimizer, an initial learning rate of 0.04 with batch size 32 due to larger input image sizes, and train for 430 and 200 epochs, respectively, following \cite{HR-NAS}.
For COCO keypoint, we follow \cite{wang2020deep} to use an Adam optimizer for 210 epochs, the initial learning rate is set to 1e-3, and is divided by 10 at the 170th and 200th epochs, respectively.
In addition, we perform architecture search during supernet training. For all search units, we use the scales from their attached BN layers as importance factors $r$; search units with $r < 0.001$ are regarded as unimportant and removed every 10 epochs (i.e., $t_s = 10$); Correspondingly, magnitude-based pruning will be performed per 25 epochs for ImageNet and Cityscapes, or per 15 epochs for ADE20K and COCO keypoint (i.e., $t_p = 25/15$), resulting intervals for sparse training as in Sec. \ref{sec:identifier}.

\textbf{Baselines and Evaluation Metrics.}
\underline{Baselines.}
For all experiments, we consider the S+P pipeline as one of our baselines, where the search method follows \cite{HR-NAS}; the pruning methods can be chosen from random pruning, magnitude pruning \cite{han2015deep,frankle2018the}, and gradient pruning \cite{lee2018snip}.
In addition, we also benchmark with hand-crafted DNNs, e.g., ShuffleNet \cite{zhang2018shufflenet,ma2018shufflenet} and MobiletNetV2 \cite{sandler2018mobilenetv2}, and prior typical NAS resulting task-specific DNNs, e.g., MobileNetV3 \cite{howard2019searching} and Auto-DeepLab \cite{liu2019auto}.
%
\hr{We do not compare with NAS/tickets works with SOTA accuracy due to different goals and experimental settings. All baselines are benchmarked under similar FLOPs or accuracy for fair comparisons.}
\underline{Evaluation Metrics.}
We evaluate the SuperTickets and all baselines in terms of accuracy-efficiency trade-offs.
Specifically, the accuracy metrics refer to top-1/5 accuracy for classification tasks; mIoU, mAcc, and aAcc for segmentation tasks; AP, AR, AP$^M$, and AP$^L$ for human pose estimation tasks. For efficiency metrics, we evaluate and compare both the number of parameters and inference FLOPs.

\subsection{Evaluating SuperTickets over Typical Baselines}

\subsubsection{SuperTickets on the Classification Task}
\ 
\newline

\begin{figure}[t]
    \centering
    \includegraphics[width=0.9\linewidth]{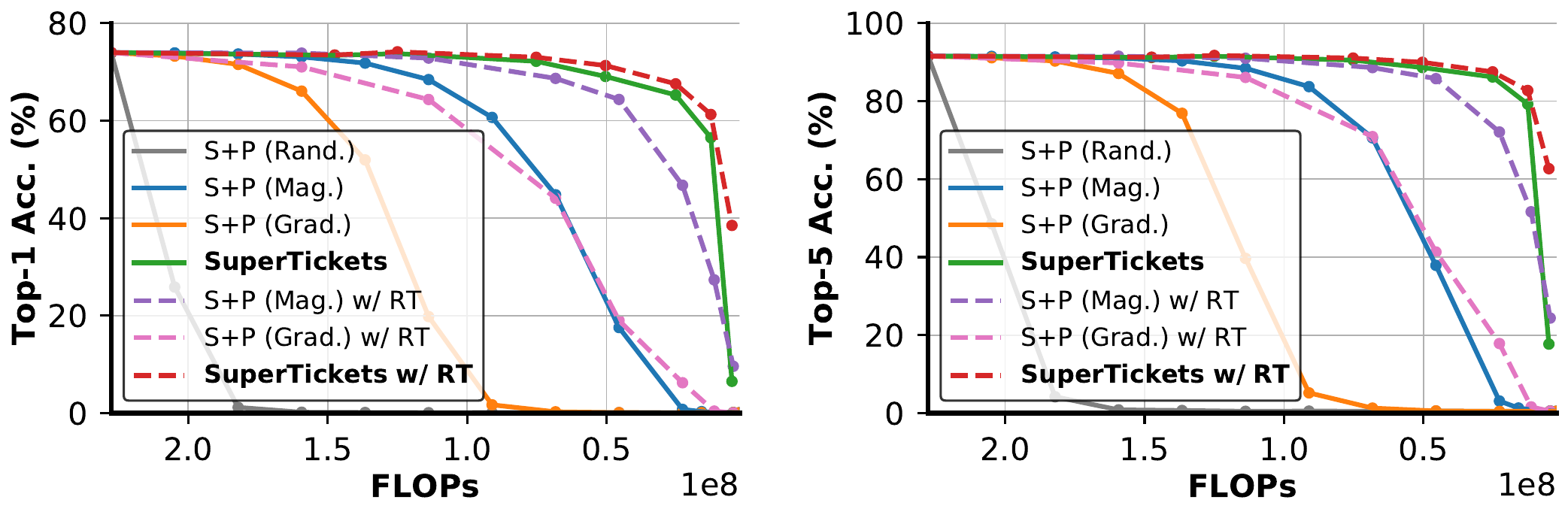}
    \caption{Comparing the top-1/5 accuracy and FLOPs of the proposed SuperTickets and S+P baselines on ImageNet. Each method has a series of points to represent different pruning ratios ranging from 10\% to 98\%. All accuracies are averaged over three runs. We also benchmark all methods with retraining (denoted as w/ RT).}
    \label{fig:imagenet}
\end{figure}

\begin{wraptable}{r}{0.55\textwidth}
    \setlength{\tabcolsep}{4pt}
    \centering
    \caption{SuperTickets vs. \bp{some typical methods} on ImageNet. FLOPs is measured with the input size of 224$\times$224.}
    \resizebox{0.55\textwidth}{!}{
        \begin{tabular}{p{0.25\textwidth}|ccc}
        \hline
        \textbf{Model} & \textbf{Params} & \textbf{FLOPs} & \textbf{Top-1 Acc.} \\
        \hline
        \hline
        CondenseNet~\cite{huang2018condensenet} & 2.9M & 274M & 71.0\%  \\
        ShuffleNetV1~\cite{zhang2018shufflenet} & 3.4M & 292M & 71.5\% \\
        ShuffleNetV2~\cite{ma2018shufflenet} & 3.5M & 299M & 72.6\% \\
        MobileNetV2~\cite{sandler2018mobilenetv2} & 3.4M & 300M & 72.0\% \\
        FBNet~\cite{wu2019fbnet} & 4.5M & 295M & 74.1\%  \\
        \hline
        S+P (Grad.) & 2.7M & 114M & 64.3\%  \\
        S+P (Mag.) & 2.7M & 114M & 72.8\%  \\
        \hline
        \textbf{SuperTickets} & 2.7M & 125M & \textbf{74.2\%} \\
        \hline
        \end{tabular}%
    }
    \label{tab:imagenet}
\end{wraptable}
\noindent We show the overall comparisons between SuperTickets and \bp{some typical} baselines in terms of accuracy-efficiency trade-offs in Fig. \ref{fig:imagenet} and Table. \ref{tab:imagenet}, from which we have \textbf{two observations}. 
\underline{First}, SuperTickets consistently outperform all baselines by reducing the inference FLOPs while achieving a comparable or even better accuracy. Specifically, SuperTickets reduce 61.4\% $\sim$ 81.5\% FLOPs while offering a comparable or better accuracy (+0.1\% $\sim$ +4.6\%) as compared to both S+P and \bp{some task-specific} DNNs; Likewise, when comparing under comparable number of parameters or FLOPs, SuperTickets lead to 
on average 26.5\% (up to 64.5\%) and
on average 41.3\% (up to 71.9\%)
top-1 accuracy improvements as compared to S+P (Mag.) and S+P (Grad.) across various pruning ratios, e.g., under 50\% pruning ratios, SuperTickets achieve 74.2\% top-1 accuracy, +1.4\% and +9.9\% over S+P (Mag.) and S+P (Grad.), respectively.
\underline{Second}, SuperTickets w/o retraining even surpass S+P baselines with retraining as demonstrated in Fig. \ref{fig:imagenet}, leading to on average 6.7\% (up to 29.2\%) higher top-1 accuracy under comparable FLOPs across various pruning ratios (10\% $\sim$ 98\%). Furthermore, SuperTickets w/ retraining achieve 0.1\% $\sim$ 31.9\% (on average 5.3\%) higher accuracy than the counterparts w/o retraining, pushing forward the frontier of 
\begin{wraptable}{r}{0.47\textwidth}
    \setlength{\tabcolsep}{4pt}
    \centering
    \caption{SuperTickets vs. \bp{some typical methods} on Cityscapes. FLOPs is measured with the input size of 512$\times$1024.}
    \resizebox{0.47\textwidth}{!}{
        \begin{tabular}{p{0.25\textwidth}|ccc}
        \hline
        \textbf{Model} & \textbf{Params} & \textbf{FLOPs} & \textbf{mIoU} \\
        \hline
        \hline
        BiSeNet~\cite{yu2018bisenet} & 5.8M & 6.6G & 69.00\%  \\
        MobileNetV3~\cite{howard2019searching} & 1.5M & 2.5G & 72.36\% \\
        ShuffleNetV2~\cite{ma2018shufflenet} & 3.0M & 6.9G & 71.30\% \\
        Auto-DeepLab~\cite{liu2019auto} & 3.2M & 27.3G & 71.21\% \\
        SqueezeNAS~\cite{shaw2019squeezenas} & 0.73M & 8.4G & 72.40\%  \\
        \hline
        S+P (Grad.) w/ RT & 0.63M & 1.0G & 60.66\%  \\
        S+P (Mag.) w/ RT & 0.63M & 1.0G & 72.31\%  \\
        \hline
        \textbf{SuperTickets} & 0.63M & 1.0G & \textbf{72.68\%} \\
        \hline
        \end{tabular}%
    }
    \label{tab:cityscapes}
\end{wraptable}
accuracy-efficiency trade-offs.

\subsubsection{SuperTickets on the Segmentation Task}
\ 
\newline


\begin{figure}[t]
    \centering
    \includegraphics[width=\linewidth]{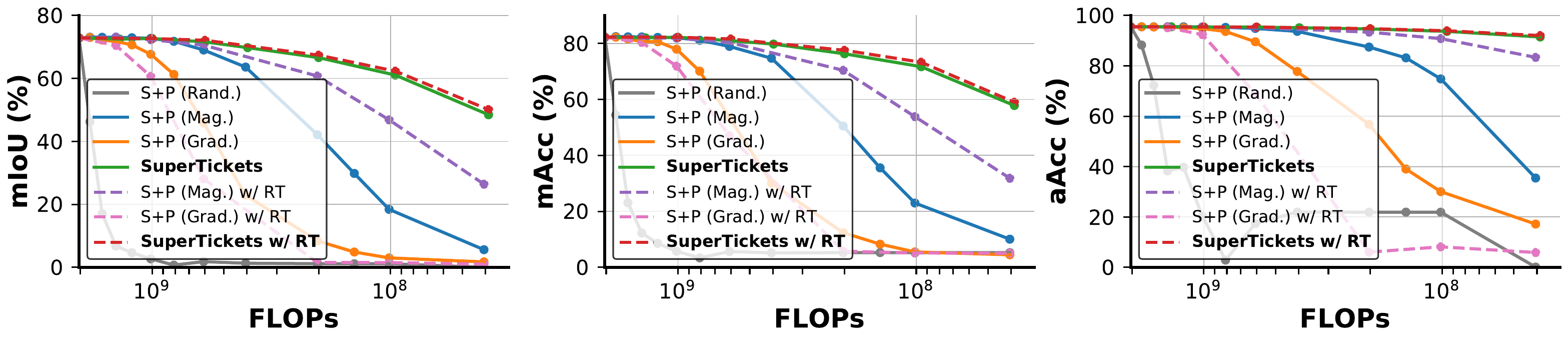}
    \centerline{(a) Comparing SuperTickets with S+P baselines on Cityscapes.}
    \includegraphics[width=\linewidth]{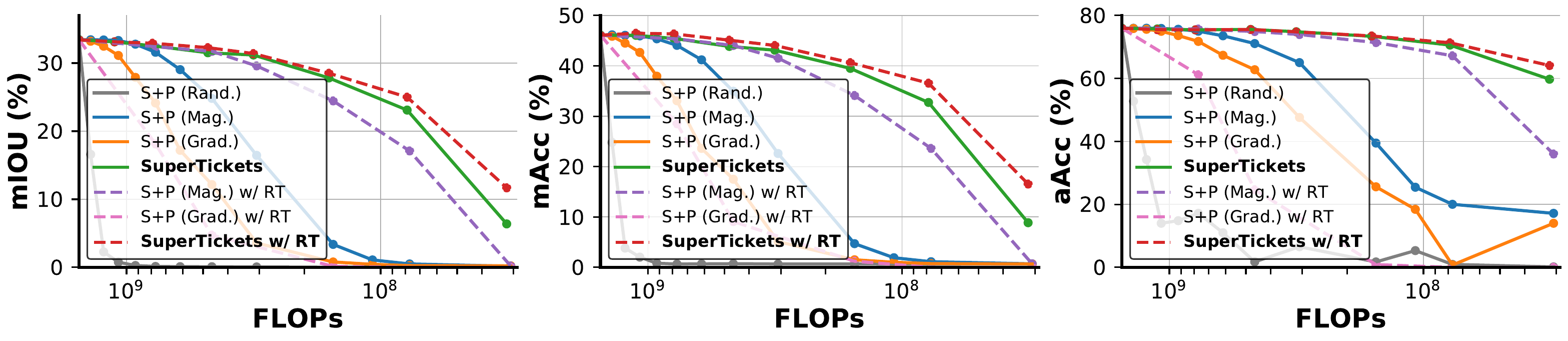}
    \centerline{(b) Comparing SuperTickets with S+P baselines on ADE20K.}
    \caption{Comparing the mIoU, mAcc, aAcc and inference FLOPs of the proposed SuperTickets and S+P baselines on Cityscapes and ADE20K datasets. Each method has a series of points to represent different pruning ratios ranging from 10\% to 98\%.}
    \label{fig:cityscapes_and_ade20k}
\end{figure}

\noindent \textbf{Experiments on Cityscapes.}
We compare SuperTickets with \bp{typical baselines} on Cityscapes as shown in Fig. \ref{fig:cityscapes_and_ade20k} (a) and Table \ref{tab:cityscapes}. We see that SuperTickets consistently outperform all baselines in terms of mIoU/mAcc/aAcc and FLOPs.
Specifically, SuperTickets reduce 60\% $\sim$ 80.86\% FLOPs while offering a comparable or better mIoU (0.28 \% $\sim$ 43.26\%) as compared to both S+P and task-specific DNNs; Likewise, when comparing under comparable number of parameters or FLOPs, SuperTickets lead to on average 
17.70\% (up to 42.86\%)
and 
33.36\% (up to 58.05\%)
mIoU improvements as compared to S+P (Mag.) and S+P (Grad.) across various pruning ratios, e.g., under 50\% pruning ratios, SuperTickets achieve 72.68\% mIoU, +0.37\% and +12\% over S+P (Mag.) and S+P (Grad.), respectively.
We also report the comparison among methods after retraining at Fig. \ref{fig:cityscapes_and_ade20k}, as denoted by ``w/ RT''. 
We find that S+P (Grad.) w/ RT suffers from overfitting and even leads to worse performance;
In contrast, SuperTickets w/ retraining further achieve 0.51\% $\sim$ 1.64\% higher accuracy than the counterparts w/o retraining, pushing forward the frontier of accuracy-efficiency trade-offs.

\begin{wraptable}{r}{0.55\textwidth}
    \setlength{\tabcolsep}{4pt}
    \centering
    \caption{SuperTickets vs. \bp{typical methods} on ADE20K. FLOPs is measured with the input size of 512$\times$512.}
    \resizebox{0.55\textwidth}{!}{
        \begin{tabular}{p{0.25\textwidth}|ccc}
        \hline
        \textbf{Model} & \textbf{Params} & \textbf{FLOPs} & \textbf{mIoU} \\
        \hline
        \hline
        MobileNetV2~\cite{sandler2018mobilenetv2} & 2.2M & 2.8G & 32.04\% \\
        MobileNetV3~\cite{howard2019searching} & 1.6M & 1.3G & 32.31\% \\
        \hline
        S+P (Grad.) & 1.0M & 0.8G & 24.14\% \\
        S+P (Mag.) & 1.0M & 0.8G & 31.59\% \\
        \hline
        \textbf{SuperTickets} & 1.0M & 0.8G & \textbf{32.54\%} \\
        \hline
        \end{tabular}%
    }
    \label{tab:ade20k}
\end{wraptable}

\textbf{Experiments on ADE20K.}
Similarly, we test the superiority of SuperTickets on ADE20K as shown in Fig. \ref{fig:cityscapes_and_ade20k} (b) and Table \ref{tab:ade20k}.
The proposed SuperTickets consistently outperform all baselines in terms of accuracy-efficiency trade-offs, reducing 38.46\% $\sim$ 48.53\% FLOPs when comparing under similar mIoU.
When compared under comparable number of parameters or FLOPs, 
SuperTickets lead to an average of
9.43\% (up to 22.6\%)
and 
14.17\% (up to 27.61\%)
mIoU improvements as compared to S+P (Mag.) and S+P (Grad.), respectively, across various pruning ratios.
In addition, SuperTickets w/ retraining further achieve 0.01\% $\sim$ 5.3\% higher accuracy than the counterparts w/o retraining on ADE20K.

\subsubsection{SuperTickets on the Human Pose Estimation Task}
\ 
\newline

\begin{figure}[t]
    \centering
    \includegraphics[width=\linewidth]{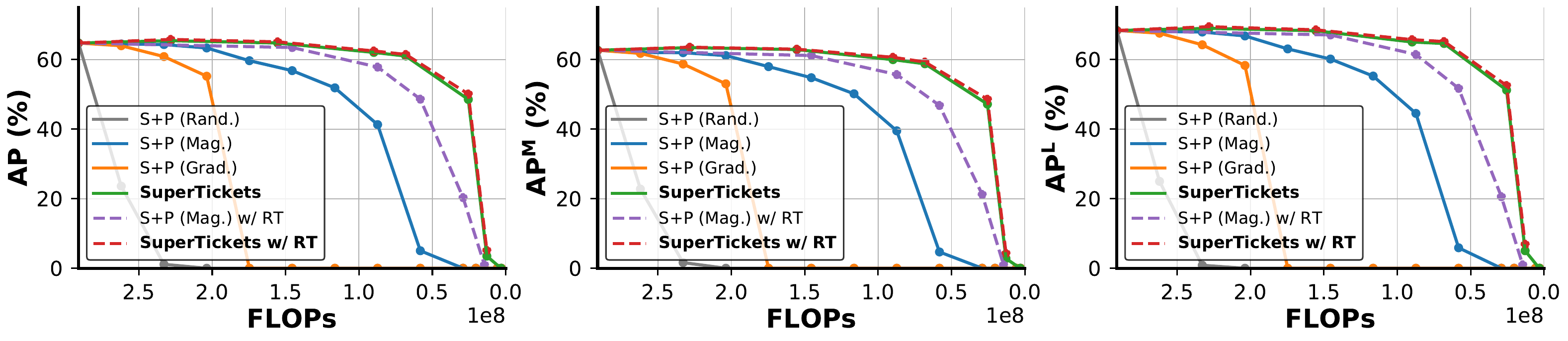}
    \caption{Comparing the AP, AP$^M$, AP$^L$ and inference FLOPs of the proposed SuperTickets and baselines on human pose estimation task and COCO keypoint dataset. Each method has a series of points for representing different pruning ratios ranging from 10\% to 98\%. All accuracies are averaged over three runs.}
    \label{fig:coco}
\end{figure}

\noindent We compare SuperTickets with \bp{a few typical} baselines on COCO keypoint as shown in Fig. \ref{fig:coco} and Table \ref{tab:coco}. We see that SuperTickets consistently outperform 
\begin{wraptable}{r}{0.55\textwidth}
    \setlength{\tabcolsep}{3pt}
    \centering
    \caption{SuperTickets vs. \bp{typical algorithms} on COCO. FLOPs is measured with the input size of 256$\times$192.}
    \resizebox{0.55\textwidth}{!}{
        \begin{tabular}{p{0.22\textwidth}|cc|cccc}
        \hline
        \textbf{Model} & \textbf{Params} & \textbf{FLOPs} & \textbf{AP} & \textbf{AP$^M$} & \textbf{AP$^L$} & \textbf{AR} \\
        \hline
        \hline
        ShuffleNetV1~\cite{zhang2018shufflenet} & 1.0M & 0.16G & 58.5 & 55.2 & 64.6 & 65.1 \\
        ShuffleNetV2~\cite{ma2018shufflenet} & 1.3M & 0.17G & 59.8 & 56.5 & 66.2 & 66.4 \\
        MobileNetV2~\cite{sandler2018mobilenetv2} & 2.3M & 0.33G & 64.6 & 61.0 & 71.1 & 70.7 \\
        S+P (Mag.) & 0.6M & 0.23G & 63.4 & 61.2 & 66.8 & 67.3 \\
        \hline
        \textbf{SuperTickets} & 0.6M & 0.23G & \textbf{65.4} & \textbf{63.4} & 69.0 & 68.9 \\
        \hline
        \end{tabular}%
    }
    \label{tab:coco}
\end{wraptable}
all \bp{the related} baselines in terms of AP/AP$^M$/AP$^L$/AR and FLOPs. Specifically, SuperTickets reduce 30.3\% $\sim$ 78.1\% FLOPs while offering a comparable or better AP (+0.8\% $\sim$ 11.79\%) as compared to both S+P and task-specific DNNs;
Likewise, when comparing under comparable number of parameters or FLOPs, SuperTickets lead to on average 17.4\% (up to 55.9\%) AP improvements. 
In addition, SuperTickets w/ retraining further achieve on average 1.1\% higher accuracy than the counterparts w/o retraining on COCO keypoint.

\subsection{Ablation Studies of the Proposed SuperTickets}

\subsubsection{Ablation Studies of SuperTickets' Identification}
\ 
\newline

\begin{figure}[t]
    \centering
    \includegraphics[width=\linewidth]{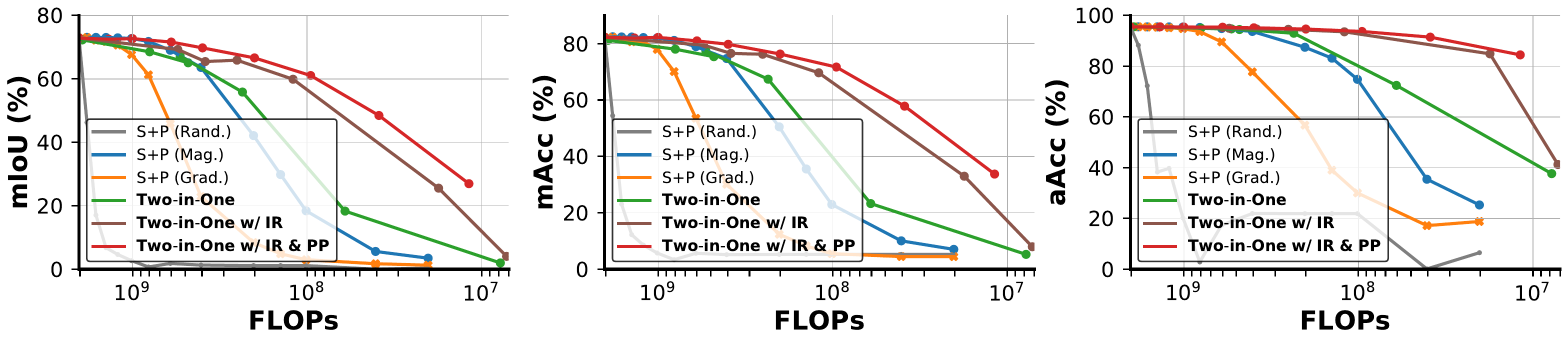}
    \caption{Ablation studies of the SuperTickets identified from two-in-one framework w/ or w/o the proposed iterative activation (IR) and progressive pruning (PP) techniques.}
    \label{fig:cityscapes_ablation}
\end{figure}

\noindent We provide comprehensive ablation studies to show the benefit breakdown of the proposed two-in-one training framework and more effective identification techniques, i.e., progressive pruning (PP) and iterative reactivation (IR). As shown in Fig. \ref{fig:cityscapes_ablation}, we report the complete mIoU-FLOPs trade-offs with various pruning ratios ranging from 10\% to \bp{99\%} when testing on Cityscapes dataset, where x axis is represented by log-scale for emphasizing the improvements when pruning ratio reaches high.
As compared to S+P (Mag.), SuperTickets identified from vanilla two-in-one framework achieve up to 40.17\% FLOPs reductions when comparing under similar mIoU, or up to 13.72\% accuracy improvements when comparing under similar FLOPs;
Adopting IR during two-in-one training further leads to up to 68.32\% FLOPs reductions or up to 39.12\% mIoU improvements;
On top of the above, adopting both IR and PP during two-in-one training offers up to 80.86\% FLOPs reductions or up to 43.26\% mIoU improvements.
This set of experiments validate the effectiveness of the general two-in-one framework and each of the proposed techniques.

\subsubsection{Ablation Studies of SuperTickets' Transferability} \label{sec:transfer_ablation}
\ 
\newline

\begin{figure}[t]
    \centering
    \includegraphics[width=\linewidth]{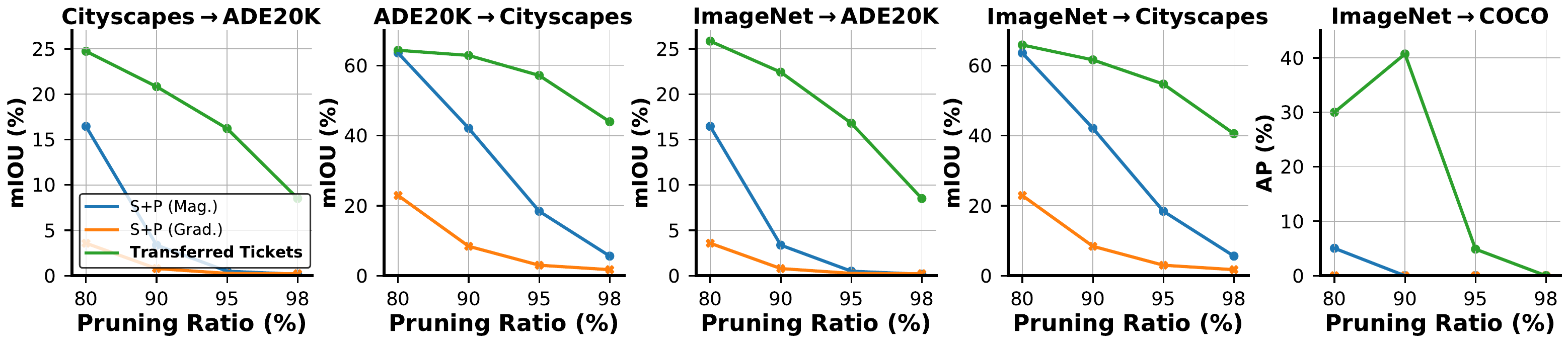}
    \caption{Ablation studies of transferring identified SuperTickets from one dataset/task to another dataset/task under various pruning ratios ranging from 80\% to 98\%.}
    \label{fig:transfer_ablation}
\end{figure}

\noindent We previously use one set of experiments under 90\% sparsity in Sec. \ref{sec:transfer} to validate that the identified SuperTickets can transfer well. In this section, we supply more comprehensive ablation experiments under various pruning ratios and among several datasets/tasks. As shown in Fig. \ref{fig:transfer_ablation}, 
the left two subplots indicate the transfer between different datasets (Cityscapes $\leftrightarrow$ ADE20K) generally works across four pruning ratios. In particular, transferred SuperTickets lead to 76.14\% $\sim$ 81.35\% FLOPs reductions as compared to the most competitive S+P baseline, while offering comparable mIoU (0.27\% $\sim$ 1.85\%).
Furthermore, the right three subplots validate that the identified SuperTickets from classification task can transfer well to other tasks (i.e., segmentation and human pose estimation). Specifically, it leads to 68.67\% $\sim$ 69.43\% FLOPs reductions as compared to the S+P (Mag.) baseline, when achieving comparable mIoU or AP.

%% file: sections/4-Conclusion.tex
\section{Conclusion}

In this paper, we advocate a two-in-one framework where both efficient DNN architectures and their lottery subnetworks (i.e., SuperTickets) can be identified from a supernet simultaneously, resulting in better performance than first-search-then-prune baselines.
Also, we develop two techniques during supernet training to more effectively identify such SuperTickets, pushing forward the frontier of accuracy-efficiency trade-offs.
Moreover, we test the transferability of SuperTickets to reveal their potential for
being task-agnostic.
Results on three tasks and four datasets consistently demonstrate the superiority of proposed two-in-one framework and the resulting SuperTickets, opening up a new perspective in searching and pruning for more accurate and efficient networks.

\section*{Acknowledgement}
We would like to acknowledge the funding support from the NSF NeTS funding (Award number: 1801865) and NSF SCH funding (Award number: 1838873) for this project.

%% file: sections/5-Appendix.tex
\newpage
\appendix

\section{Visualization of The Adopted Supernet Architecture}

\begin{figure}[h]
    \centering
    \includegraphics[width=\linewidth]{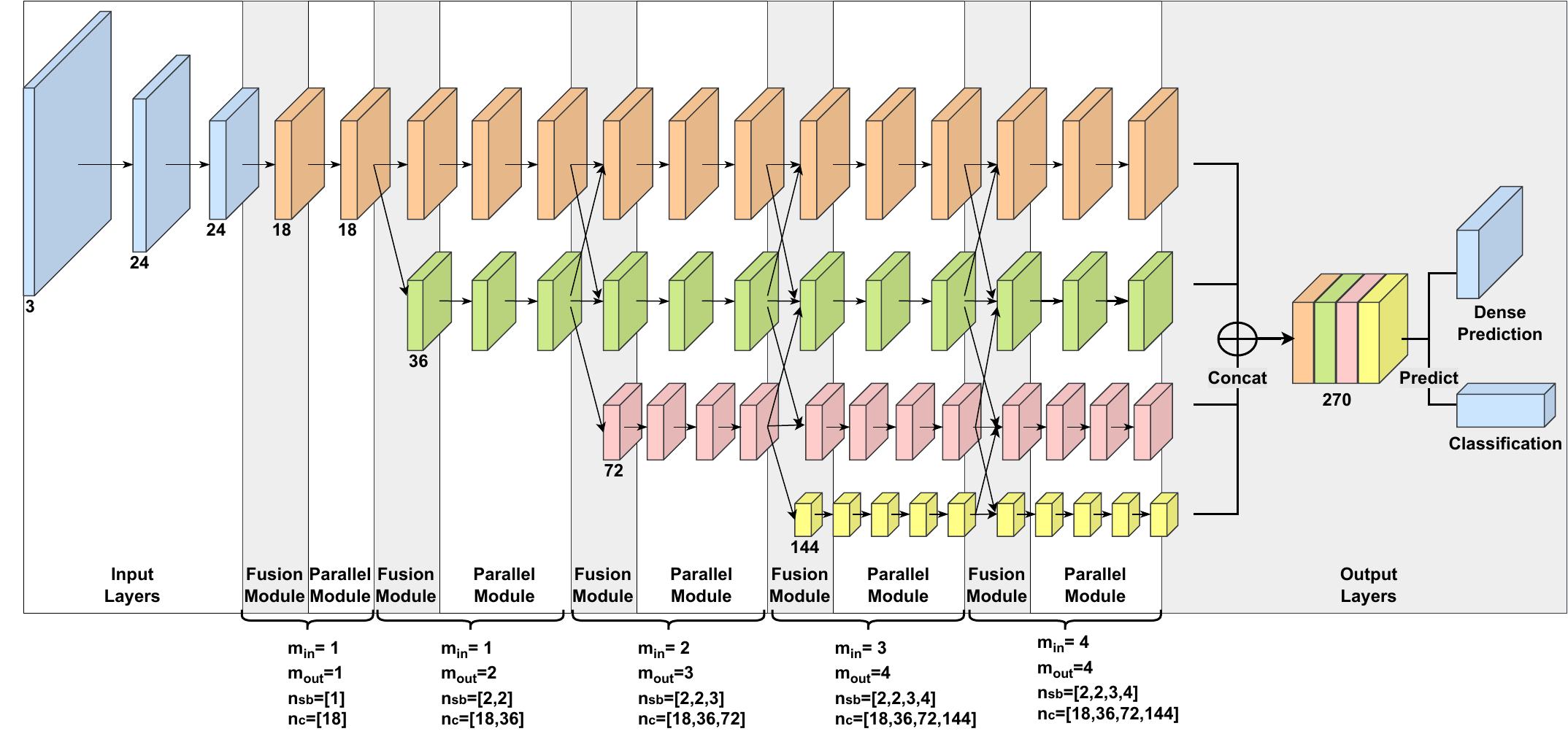}
    \caption{Visualization of the adopted supernet architecture, where $m_{in}$ and $m_{out}$ denote the number of input and output branches in the fusion module; $n_{sb}$ and $n_c$ represent the number of searching blocks and channels in the parallel module, respectively.}
    \label{fig:supernet}
\end{figure}

We visualize the adopted supernet following \cite{HR-NAS} in Fig. \ref{fig:supernet}. It begins with two 3$\times$3 convolutions with stride 2, which are followed by five fusion modules and five parallel modules to gradually divide it into four branches of decreasing resolutions, the learned features from all branches are then merged together for classification or dense prediction.

\section{SuperTickets (ST) vs. Random Pruning (RP) and Random Re-Initialization (RR-Init).}
We compare the proposed SuperTickets (ST) with both the ``ST w/ RP'' and ``ST w/ RR-Init'' in Table \ref{tab:compare_rp}.
We consider two datasets under 80\% and 90\% sparsity: ST consistently outperforms the two baselines, achieving on-average \hr{36.28\%/42.03\%/21.95\% and 11.20\%/12.27\%/4.34\%} mIoU/mAcc/aAcc improvements over ``ST w/ RP'' and ``ST w/ RR-Init'', respectively, under a comparable number of parameters and FLOPs. These experiments show that SuperTickets performs better than both RP and RR-Init, which is consistent with the LTH finding.

\begin{table}[h]
    \centering
    \caption{Comparing ST with ``ST w/ RP'' and ``ST w/ RR-Init'' on both Cityscapes and ADE20K under 80\% and 90\% sparsity.}
    \resizebox{\linewidth}{!}{
    \begin{tabular}{l|c|ccc||c|ccc}
    \hline
    \multicolumn{1}{c|}{\multirow{2}[4]{*}{\textbf{Methods}}} &
    \multirow{2}[4]{*}{\textbf{FLOPs}} & \multicolumn{3}{c||}{\textbf{Cityscapes ($p$ = 80\%)}} &
    \multirow{2}[4]{*}{\textbf{FLOPs}} &
    \multicolumn{3}{c}{\textbf{Cityscapes ($p$ = 90\%)}} \\
    \cline{3-5} \cline{7-9}  &  & \textbf{mIoU} & \textbf{mAcc} & \textbf{aAcc} & & \textbf{mIoU} & \textbf{mAcc} & \textbf{aAcc} \\
    \hline
    S+P w/ RP & 405M & 1.30\% & 5.17\% & 21.96\% & 203M & 1.15\% & 5.26\% & 21.9\% \\
    ST w/ RP & 397M & 20.17\% & 27.57\% & 68.73\% & 200M & 16.55\% & 23.83\% & 65.90\% \\
    ST w/ RR-Init & 397M & 56.88\% & 67.33\% & 92.62\% & 200M & 52.96\%  & 62.66\% & 91.91\% \\
    \textbf{ST} & 397M & \textbf{69.77\%} & \textbf{79.76\%} & \textbf{95.12\%} & 200M & \textbf{66.61\%} & \textbf{76.30\%} & \textbf{94.63\%} \\
    \hline
    \hline
    \multicolumn{1}{c|}{\multirow{2}[4]{*}{\textbf{Methods}}} &
    \multirow{2}[4]{*}{\textbf{FLOPs}} & \multicolumn{3}{c||}{\textbf{ ADE20K ($p$ = 80\%)}} &
    \multirow{2}[4]{*}{\textbf{FLOPs}} &
    \multicolumn{3}{c}{\textbf{ ADE20K ($p$ = 90\%)}} \\
    \cline{3-5} \cline{7-9}  &   & \textbf{mIoU} & \textbf{mAcc} & \textbf{aAcc} & & \textbf{mIoU} & \textbf{mAcc} & \textbf{aAcc} \\
    \hline
    S+P w/ RP & 308M & 0.06\% & 0.66\% & 6.54\% & 154M & 0.01\% & 0.66\% & 1.72\% \\
    ST w/ RP & 317M & 8.58\% & 11.93\% & 60.21\% & 159M & 4.98\% & 7.19\% & 55.30\% \\
    ST w/ RR-Init & 317M & 21.24\% & 30.62\% & 68.57\% & 159M & 19.49\%  & 28.46\% & 67.26\% \\
    \textbf{ST} & 317M & \textbf{31.19\%} & \textbf{43.10\%} & \textbf{74.82\%} & 159M & \textbf{27.82\%} & \textbf{39.49\%} & \textbf{73.37\%} \\
    \hline
    \end{tabular}%
    }
    \label{tab:compare_rp}
\end{table}

\textbf{Transferability of ST vs. RP and RR-Init.}
Similarly, we compare the transferability of the three ST variants when transferring them across different datasets, including (1) ADE20K $\rightarrow$ Cityscapes or (2) Cityscapes $\rightarrow$ ADE20K. As shown in Table \ref{tab:compare_rp_transfer}, ST achieves on-average \hr{33.14\%/39.38\%/21.92\% and 11.37\%/13. 67\%/3.20\%} mIoU/mAcc/aAcc improvements over the ``ST w/ RP'' and ``ST w/ RR-Init'' baselines, respectively, indicating that RP and RR-Init are inferior in transferability as compared to the proposed ST.

\begin{table}[t]
    \centering
    \caption{ST variants transfer validation tests under 90\% sparsity.}
    \resizebox{\linewidth}{!}{
    \begin{tabular}{l|ccc||l|ccc}
    \hline
    \multicolumn{1}{c|}{\multirow{2}[4]{*}{\textbf{Methods}}} & \multicolumn{3}{c||}{\textbf{ADE20K $\rightarrow$ Cityscapes}} & \multicolumn{1}{c|}{\multirow{2}[4]{*}{\textbf{Methods}}} & \multicolumn{3}{c}{\textbf{Cityscapes $\rightarrow$ ADE20K}}  \\
    \cline{2-4}\cline{6-8}  & \textbf{mIoU} & \textbf{mAcc} & \textbf{aAcc} &   & \textbf{mIoU} & \textbf{mAcc} & \textbf{aAcc}  \\
    \hline
    ST w/ RP & 10.51\% & 14.42\% & 61.28\% & ST w/ RP & 6.95\% & 10.1\% & 57.7\% \\
    ST w/ RR-Init & 46.19\% & 54.92\% & 90.88\% & ST w/ RR-Init & 14.82\% & 21.02\% & 65.54\% \\
    ST & \textbf{62.91\%} & \textbf{73.32\%} & \textbf{93.82\%} & ST & \textbf{20.83\%} & \textbf{29.95\%} & \textbf{69.00\%}  \\
    \hline
    \end{tabular}%
    }
    \label{tab:compare_rp_transfer}
\end{table}

\section{Clarification of the LTH Settings.}
There are two confusing settings when talking about LTH: (1) directly test the accuracy of the found structure and the trained weights; and (2) the weights are restored to their initial value and trained with the obtained mask to obtain test accuracy.
We tried both of the aforementioned settings and find the former, i.e., directly testing the accuracy of the found structure and trained weights, has already achieved good results. This is another highlight of our work, as it can help to largely save the retraining time.
Furthermore, to address your concern, we re-initialize the SuperTickets to (1) their initial values, following the origin LTH (``ST w/ LT-Init'') and (2) \textit{early} or (3) \textit{late} stages following \cite{frankle2020linear} (``ST w/ ELT-Init or LLT-Init''), and compare them with the RR-Init counterparts.
From Table \ref{tab:compare_lth}, we can see that \textbf{(1)} ST under all LTH settings achieves better accuracy than RR-Init, indicating the effectiveness of ST; \textbf{(2)} vanilla LT-Init underperforms both ELT-Init and LLT-Init under ST settings, consistent with \cite{frankle2020linear}; and \textbf{(3)} ST w/ ELT-Init or LLT-Init achieves comparable or slightly better accuracy than ST w/o Retrain at a cost of retraining.

\begin{table}[!t]
    \centering
    \caption{Comparing ST w/ various LTH settings (90\% sparsity).}
    \resizebox{\linewidth}{!}{
    \begin{tabular}{l|rrr||l|rrr}
    \hline
    \multicolumn{1}{c|}{\multirow{2}[4]{*}{\textbf{Methods}}} & \multicolumn{3}{c||}{\textbf{Cityscapes ($p$ = 90\%)}} & \multicolumn{1}{c|}{\multirow{2}[4]{*}{\textbf{Methods}}} & \multicolumn{3}{c}{\textbf{ADE20K ($p$ = 90\%)}}  \\
    \cline{2-4}\cline{6-8}  & \multicolumn{1}{c}{\textbf{mIoU}} & \multicolumn{1}{c}{\textbf{mAcc}} & \multicolumn{1}{c||}{\textbf{aAcc}} &   & \multicolumn{1}{c}{\textbf{mIoU}} & \multicolumn{1}{c}{\textbf{mAcc}} & \multicolumn{1}{c}{\textbf{aAcc}}  \\
    \hline
    ST w/ RR-Init & 52.96\% & 62.66\% & 91.91\% & ST w/ RR-Init & 19.49\% & 28.46\% &  67.26\% \\
    ST w/ LT-Init & 59.63\% & 70.24\% & 93.33\% & ST w/ LT-Init & 25.32\% & 36.76\% & 71.33\% \\
    ST w/ ELT-Init & 65.82\% & 76.74\% & 94.54\% & ST w/ ELT-Init & 25.79\% & 37.33\% & 72.10\% \\
    ST w/ LLT-Init & 67.17\% & 77.03\% & 94.73\% & ST w/ LLT-Init & 28.51\% & 40.63\% & 73.49\% \\
    ST w/o Retrain & 66.61\% & 77.03\% & 94.73\% & ST w/o Retrain & 27.82\% & 39.49\% & 73.37\% \\
    \hline
    \end{tabular}%
    }
    \label{tab:compare_lth}
\end{table}

\section{Speedups in terms of Inference Time}

In addition to the number of parameters and FLOPs, we measure the inference FPS and speedups on both 1080Ti GPUs and a SOTA sparse DNN inference accelerator \cite{qin2020sigma}.
As shown in Table \ref{tab:speedups}, ST achieves on par or even higher (i.e., \hr{1.8$\times\!\!\sim$2.9$\times$} speedups) FPS on GPUs and much reduced accelerator time (i.e., \hr{2.9$\times\!\!\sim$4.1$\times$} speedups) on \cite{qin2020sigma} than the baselines, thanks to simultaneous architecture searching and parameter pruning (i.e., 2-in-1) and ST.

\begin{table}[h]
    \centering
    \caption{ST vs. typical baselines on Cityscapes, in terms of inference time measured on both GPUs and sparse accelerators.}
    \resizebox{\linewidth}{!}{
    \begin{tabular}{l|ccc|cc}
    \hline
    \textbf{Model} & \textbf{Params} & \textbf{FLOPs} & \textbf{mIoU} & \textbf{GPU FPS} & \textbf{Sparse Acc. Time}  \\
    \hline
    BiSeNet & 5.8M & 6.6G & 69.00\% & 105.8 & 180.8ms \\
    DF1-Seg-d8 & - & - & 71.40\% & 136.9 & 181.7ms \\
    FasterSeg & 4.4M & - & 71.50\% & 163.9 & 142.4ms \\
    SqueezeNAS & 0.73M & 8.4G & 72.40\% & 117.2 & 198.5ms \\
    \hline
    \textbf{ST ($p$ = 50\%)} & 0.63M & 1.0G & \textbf{72.68\%} & 310.7 & 48.3ms \\
    \hline
    \end{tabular}%
    }
    \label{tab:speedups}
\end{table}

\section{Discussions}

\textbf{Limitations of Transferred SuperTickets.}
Although identified SuperTickets can transfer with only classifiers as task-specific, \bp{there is still} a limitation in the transferred SuperTickets. That is, transferred SuperTickets cannot surpass those SuperTickets directly found on the target datasets/tasks. Moreover, when the sparsity is low (e.g., 30\%), the transferred SuperTickets will underperform both SuperTickets and S+P.
This is counterintuitive and opposite to the observation in compressing pretrained models \cite{gordon2020compressing}, where low pruning ratios do not hurt the accuracy after transferring while overpruning leads to under-fitting.
It implies that the dedicated search is necessary when pruning ratio is relatively low; while for high sparsity, the impacts of neural architectures will be less.

\begin{figure}[t]
    \centering
    \includegraphics[width=\linewidth]{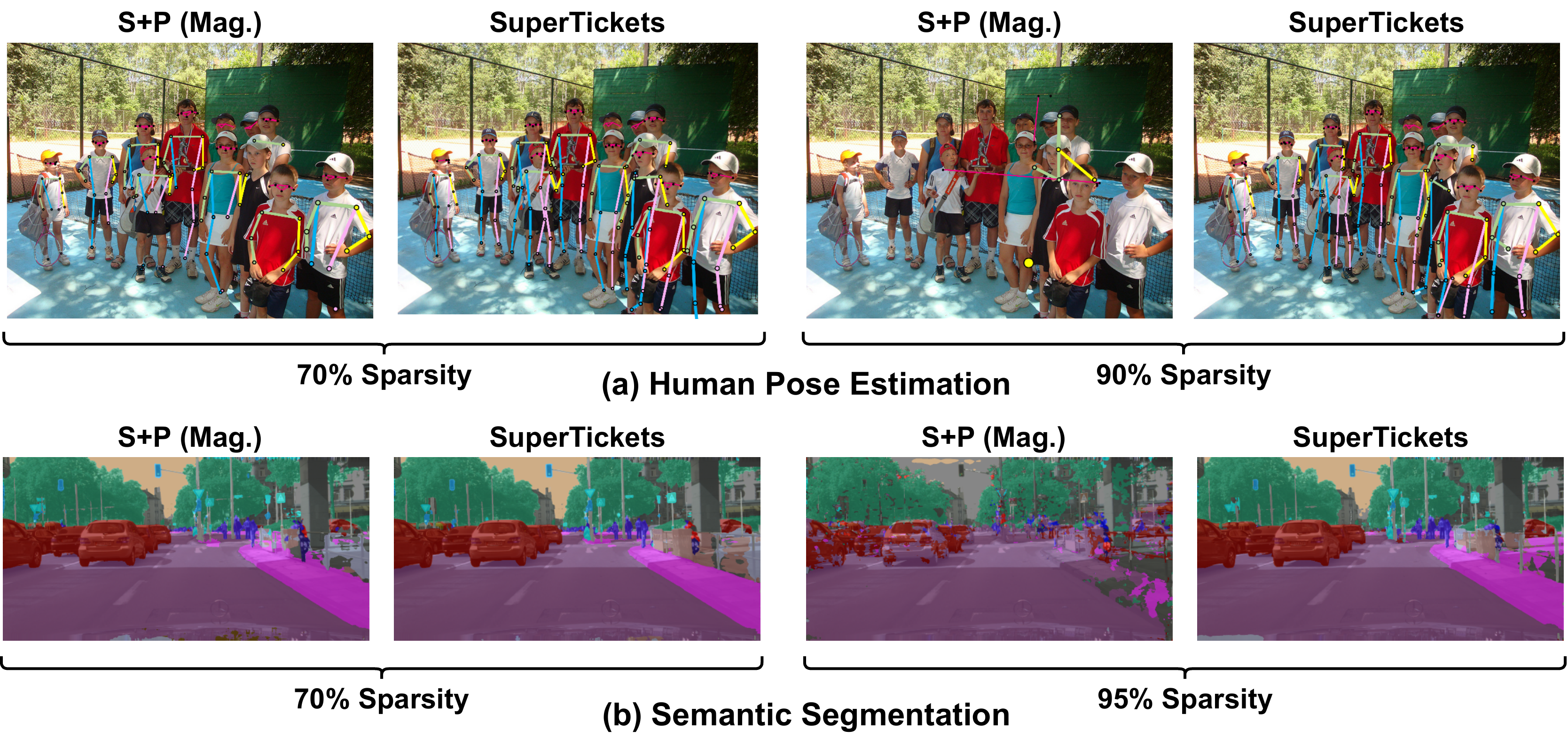}
    \caption{Visualization of the human pose estimation on COCO keypoint dataset and the streetview/semantic labels on Cityscapes dataset under different pruning ratios.
    }
    \label{fig:visualization}
\end{figure}

\textbf{Visualization and Discussion.}
We visualize the results of SuperTickets and S+P baselines on COCO keypoint and Cityscapes datasets under different pruning ratios, as shown in Fig. \ref{fig:visualization}. We observe that S+P baselines work but miss some keypoints or semantic understandings under medium sparsity (e.g., 70\%) while collapse under high pruning ratios (e.g, 90/95\%); In contrast, our identified SuperTickets consistently work well among a wide range of pruning ratios, validating the effectiveness of our proposed SuperTickets.

\section{More Visualization of Visual Recognition Results}

\begin{figure}
    \centering
    \includegraphics[width=0.95\linewidth]{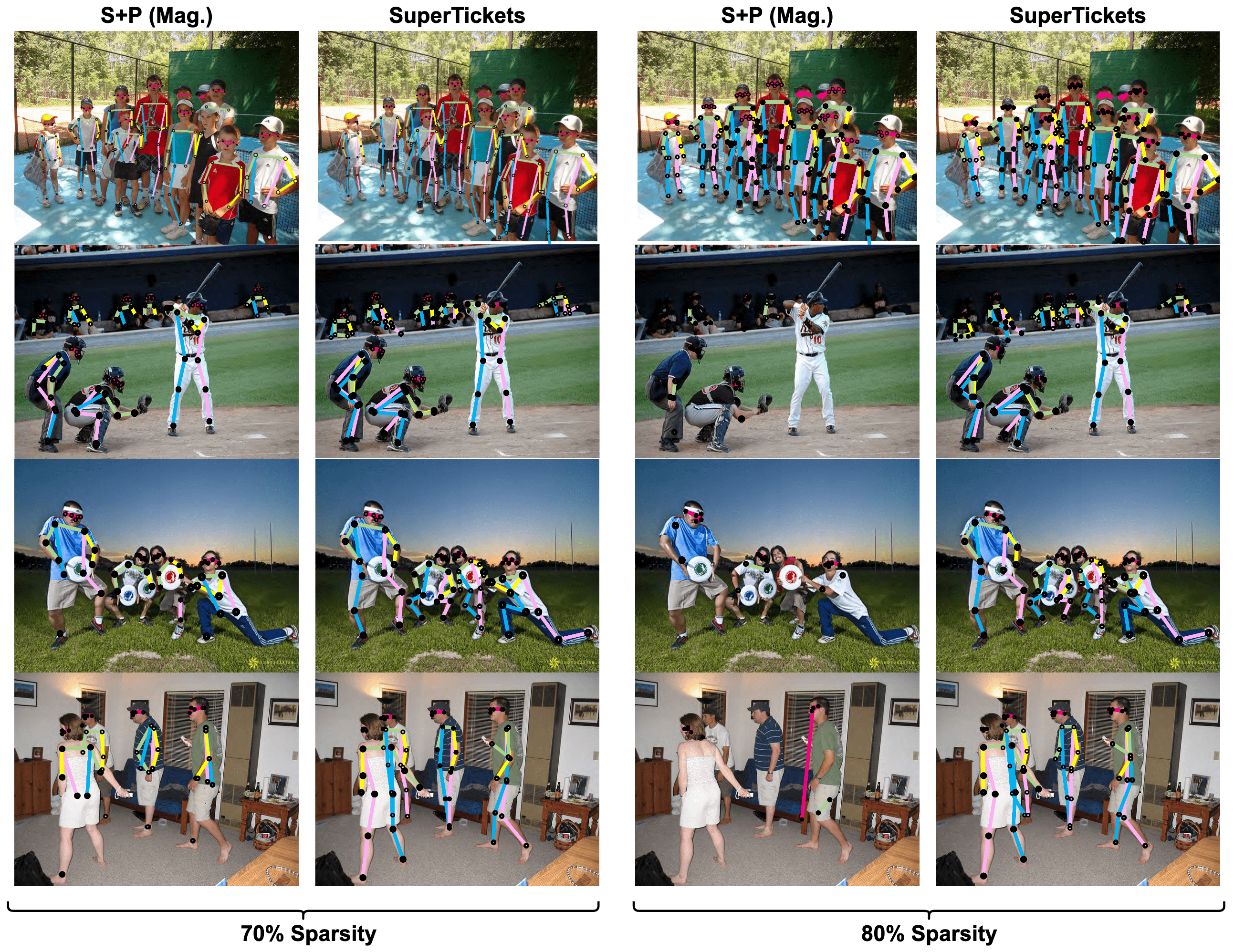}
    \includegraphics[width=0.95\linewidth]{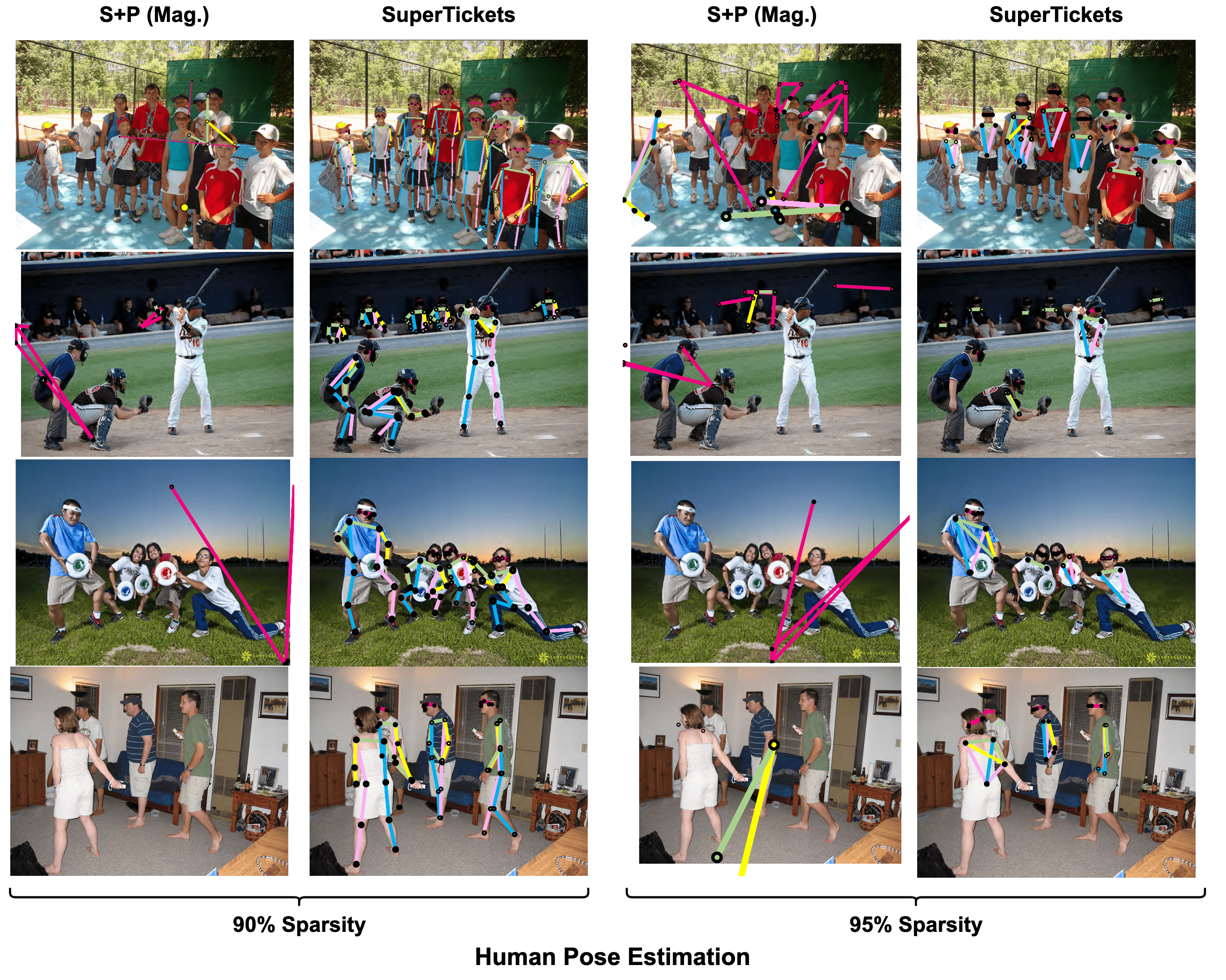}
    \caption{Visualization of the human pose estimation on COCO keypoint dataset under various pruning ratios.}
    \label{fig:COCO}
\end{figure}

\begin{figure}[t]
    \centering
    \includegraphics[width=\linewidth]{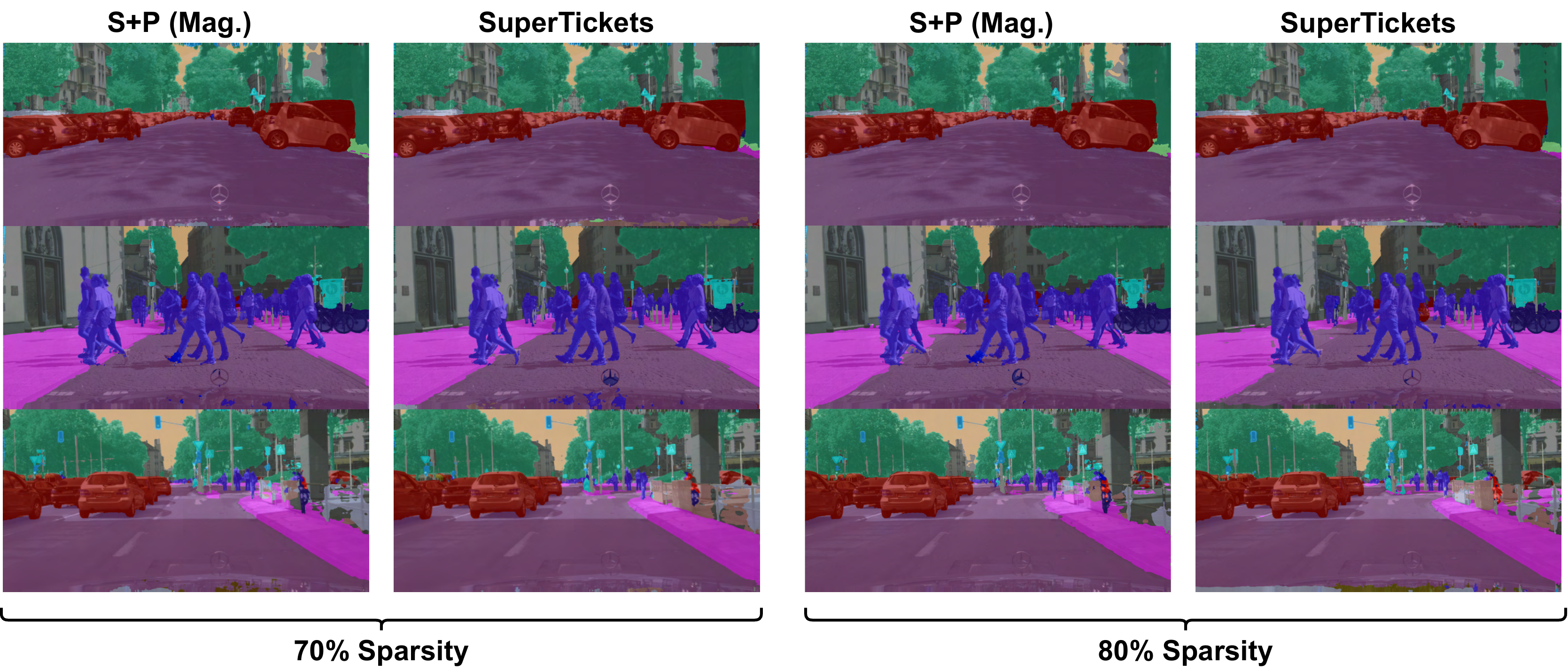}
    \includegraphics[width=\linewidth]{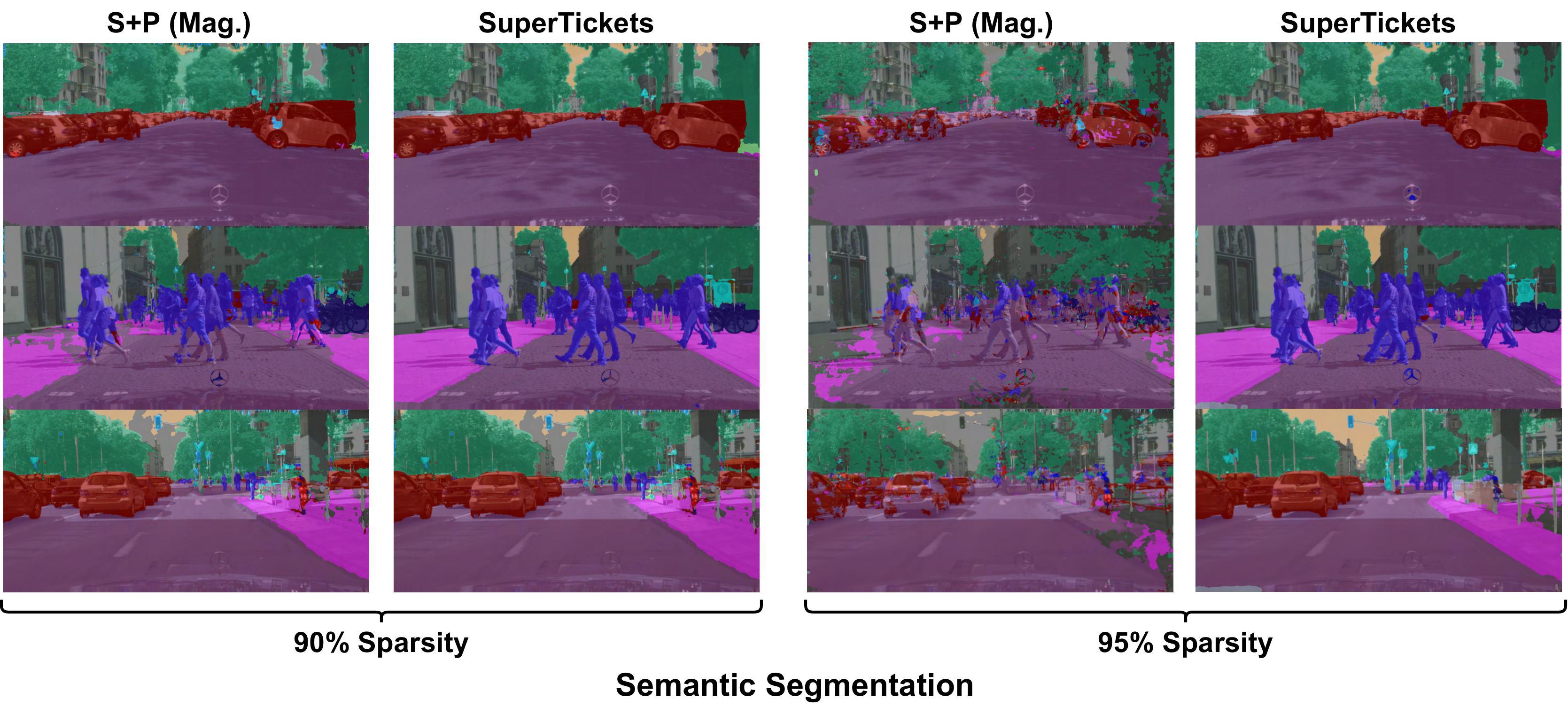}
    \caption{Visualization of the streetview/semantic labels on Cityscapes dataset under various pruning ratios.}
    \label{fig:Cityscapes}
\end{figure}

We further visualize the results of SuperTickets and S+P baselines on COCO keypoint and Cityscapes datasets under different pruning ratios, as shown in Fig. \ref{fig:COCO} and Fig. \ref{fig:Cityscapes}, respectively. We observe that S+P baselines work but miss some keypoints or semantic understandings under medium sparsity (e.g., 70/80\%) while collapse under high pruning ratios (e.g, 90/95\%); In contrast, our identified SuperTickets consistently work well among a wide range of pruning ratios, validating the effectiveness of our proposed SuperTickets.

